\pdfoutput=1

\documentclass[11pt]{article}

\usepackage{acl}
\usepackage{times}
\usepackage{latexsym}

\usepackage[T1]{fontenc}

\usepackage[utf8]{inputenc}

\usepackage{microtype}

\usepackage{inconsolata}

\usepackage{graphicx}

\usepackage{amsmath}
\usepackage{enumitem}
\setlist[enumerate]{itemsep=0pt, topsep=0pt, parsep=0pt}
\usepackage{algorithm}
\usepackage{algpseudocode}
\usepackage{algorithmicx}
\usepackage{amssymb}

\usepackage{makecell} 

\usepackage[utf8]{inputenc}
\usepackage{booktabs} 
\usepackage{rotating} 
\usepackage{url} 

\title{Arena-Lite: Efficient and Reliable Large Language Model Evaluation via Tournament-Based Direct Comparisons}

\author{
 Seonil Son\textsuperscript{1,2,§},
 Ju-Min Oh\textsuperscript{1,3,§},
 Heegon Jin\textsuperscript{1,4,§} \\
\textbf{ Cheolhun Jang\textsuperscript{1},
 Jeongbeom Jeong\textsuperscript{1},
 Kuntae Kim\textsuperscript{1,5,§}} \\
 \textsuperscript{1}NC AI, \textsuperscript{2}RLWRLD Inc., \textsuperscript{3}Samsung AI Research, \\
 \textsuperscript{4}Global AI Platform, \textsuperscript{5}Samsung Life Insurance \\
 \textsuperscript{§}Work performed primarily while at NC AI\\
 \small \textbf{Correspondence:} \href{mailto:simon.son@rlwrld.ai}{simon.son@rlwlrd.ai}
}

\begin{document}
\maketitle
\begin{abstract}

As Large Language Models (LLMs) expand across domains, LLM judges have become essential for systems evaluation. 
Current benchmarks typically compare system outputs against baselines.
This baseline-mediated approach, though convenient, yields lower reliability than direct comparison between systems.
We propose Arena-Lite which integrates tournament structure on top of head-to-head comparison.
The application of a tournament structure and direct comparison eliminates the need for baseline outputs, reduces the number of required comparisons, and allows higher reliability in system rankings.
We conducted two experiments: (1) controlled stochastic modeling and (2) empirical validation with a real LLM judge. 
Those experiments collectively demonstrate that Arena-Lite consistently achieves higher reliability with fewer comparisons, even with smaller datasets or weaker judges.
We release an easy-to-use web demonstration and code to foster adoption of Arena-Lite, streamlining model selection across research and industry communities. Arena-Lite demo and code are available on \href{https://huggingface.co/spaces/NCSOFT/ArenaLite}{\url{https://huggingface.co/spaces/NCSOFT/ArenaLite}}

\end{abstract}

\section{Introduction}

\begin{figure}
  \centering
  \includegraphics[width=\linewidth]{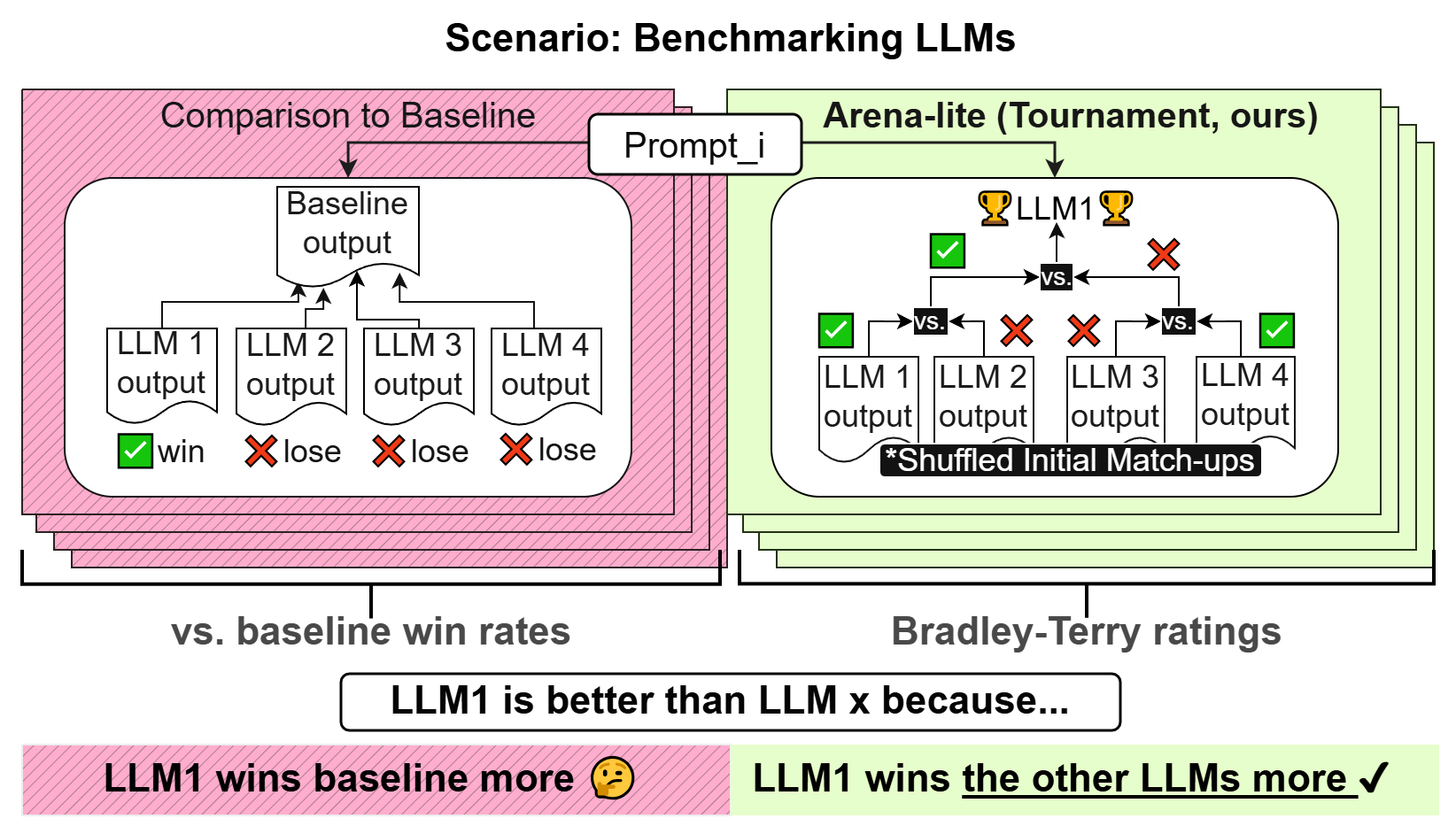}
\caption{Arena-Lite directly compares LLM response pairs over multiple single-elimination tournaments rather than comparing responses to baseline outputs. In terms of deciding whether a certain LLM is better or worse compared to the other one, we suggest direct head-to-head comparison is more intuitive and results in better separability.}
  \label{fig:va_conceptual}

\end{figure}

\begin{table}[h!]
    \centering
    \resizebox{\linewidth}{!}{
    \begin{tabular}{lll}
    \hline
      \textbf{} & \makecell[l]{Total no. \\matches ($\downarrow$)}& \makecell[l]{No. matches per\\ LLM participant ($\uparrow$)}\\ 
      \hline
      Current Practice & $n_{\text{model}} \cdot |X|$ & $|X|$ \
      \\
      Arena-Lite (ours) & $(n_{\text{model}} - 1) \cdot |X|$ & $[~|X|,~|X|* \lceil \log_2 n_\text{model} \rceil ~]$ \\
    \hline
    \end{tabular}
    }
    \caption{Comparison between Current practice of benchmarking (comparing to baseline outputs) and Arena-Lite. $|X|$ and $n_\text{model}$ represents size of benchmark dataset, and number of candidate LLMs to rank respectively. Arena-Lite, always save |X| number of comparisons for benchmarking, while allows more matches per LLM participant thanks to head-to-head comparison.}
    \label{table:comparison}
  \end{table}

LLMs excel in diverse tasks, from chatbots to code generation, due to their powerful generative capabilities~\cite{instructgpt, codellama}. 
As their versatility grows, accurately evaluating their performance becomes critical. 
To address this, benchmarks like MMLU and BigBench have emerged to assess LLM capabilities across various domains~\cite{mmlu, bigbench}. 
Many of these benchmarks, such as those for arithmetic or code execution (e.g., GSM-Hard, HumanEval~\cite{gsm-hard, humaneval}), use automated scoring to evaluate problem-solving skills. 
However, their focus is not on quality of generated content or limited to the cases where the generated contents are automatically evaluated (e.g. programming), which are mostly not the case for variety of generation tasks. 
The Chatbot Arena, a leading platform for reliable human evaluation of LLMs, has set a standard by collecting extensive human annotations~\cite{chiang2024chatbotarenaopenplatform}. 
Yet, its resource-intensive approach has prompted efforts to replicate its rankings using LLM judges as a cost-effective alternative~\cite{arena-hard, alpacaeval}. 
These methods, however, rely on baseline-mediated comparisons—comparing LLM outputs to a leading proprietary model’s outputs—which sacrifice reliability.

Current benchmarks relying on baseline often rank LLMs by their win rate against baseline responses from an leading proprietary models. 
This approach has two advantages: it scales linearly with the number of LLMs and provides a consistent quality standard. 
However, we argue that comparing LLMs directly against each other is inherently more reliable than using baseline outputs, which can introduce noise coming from weak transitivity~\cite{xu2025investigatingnontransitivityllmasajudge} of human preferences on LLM responses. 
To address this, we propose Arena-Lite, a novel evaluation framework that uses direct, head-to-head comparisons organized in a tournament structure. By eliminating the need for baseline outputs, Arena-Lite reduces the number of comparisons required while achieving stronger alignment with human-established rankings, such as those from Chatbot Arena.

Arena-Lite conducts single-elimination tournaments among participating LLMs for each prompt. From the match results, we can compute Bradley-Terry preference ratings for the final ranking~\cite{btmodel}. 
This results in a single scalar per model that captures relative performance between any model counterpart, enabling accurate and efficient ranking. 
We validate Arena-Lite through two experiments. 
The first experiment, stochastic modeling of LLM competition (Section \ref{sec:exp1}) demonstrates that tournament-based direct comparison method outperforms baseline-mediated method under various conditions, including different numbers of LLM participants, dataset rows used, and judge accuracies. Second, our empirical experiment (Section \ref{sec:exp2}) shows that Arena-Lite achieves higher correlation with Chatbot Arena’s rankings than standard approaches using baseline outputs (Table \ref{table:comparison}) attested toward number of LLM as judges. 
These results collectively highlight Arena-Lite’s ability to deliver reliable rankings with fewer comparisons, even with smaller datasets or weaker judges over various generation tasks.

\noindent Our contributions are threefold:
\begin{enumerate}
  \item We introduce Arena-Lite, a tournament-based framework for direct LLM comparisons, offering greater reliability than baseline-mediated approaches.
  \item We rigorously demonstrate, through both comprehensive modeling and empirical experiments, that Arena-Lite achieves more accurate LLM rankings while requiring fewer comparisons than prevalent practices, particularly those relying on common baseline model outputs.
  \item We open-source a functional demo and the complete code for Arena-Lite (\href{https://huggingface.co/spaces/NCSOFT/ArenaLite}{\url{https://huggingface.co/spaces/NCSOFT/ArenaLite}}), enabling researchers and industry practitioners to easily host and utilize our framework for streamlined LLM evaluation.
\end{enumerate}

\section{Preliminaries: Quantifying Generative Performance} 
\label{sec:preliminaries}

Quantifying the generative capabilities of LLMs is an inherently challenging task. 
The evaluation is complicated by the stochasticity of model outputs and the subjectivity of human judgments. 
To approximate real-world performance, a common methodology involves assessing model outputs across a diverse range of prompts. 
In this context, two metrics are widely utilized: the \textbf{win rate}, which measures the frequency of preference for a model's output over a baseline, and the \textbf{Bradley-Terry model}, which is employed to infer a latent skill rating for each model based on pairwise comparisons.

\subsection{Measuring Win rate over baseline outputs}
\label{sec:Measuring Win rate over baseline outputs}

Benchmarks like AlpacaEval and Arena-Hard-Auto assess LLM response quality by comparing it to baseline responses from proprietary model~\cite{alpacaeval, arena-hard}. An LLM judge evaluates whether the candidate LLM’s response outperforms the baseline for a given prompt. The win rate—the proportion of prompts where the LLM’s response is preferred—serves as a measure of its generative ability. While this approach is straightforward and scalable, it introduces noise coming from mediated comparisons.

\subsection{Bradley-Terry Model Preference for LLM Rating}
\label{sec:Elo Rating}

The Bradley-Terry (BT) model~\cite{btmodel} is widely used to infer baseline-mediated rankings of LLMs from pairwise comparisons. Chatbot Arena adopts the BT model rather than the classical Elo system~\cite{elo}, but both Elo and BT models are useful for expecting probability of match outcome based on a score difference, though they differ in update rules and statistical assumptions.

In the BT model, each LLM is assigned a latent score representing its procificency. 
Given LLMs $i$ and $j$ with scores $R_i$ and $R_j$, respectively, the probability that LLM $i$ is preferred over LLM $j$ is modeled as:
\begin{equation}
\text{P}(i > j) = \frac{1}{1 + 10^{(R_j - R_i)/400}}.
\label{eq:bt_model_elo}
\end{equation}
This formulation closely resembles the Elo win-probability function, reinforcing the intuitive connection between the two.

Chatbot Arena uses this BT-based formulation to rank LLMs by aggregating human preferences collected through pairwise matchups~\cite{chiang2024chatbotarenaopenplatform}. Users are shown responses from two anonymized models to the same prompt and asked to select which response they prefer. The accumulated judgments are then used to fit BT scores, producing a leaderboard that reflects relative model performance.

While this approach requires a substantial number of human evaluations to ensure reliability, it captures nuanced quality differences between models more effectively than purely automatic benchmarks. 
Arena-Lite, introduced in the next section, builds on the same BT modeling framework but seeks to reduce the number of required comparisons by using tournament-structured match-making.

\section{Arena-Lite}
\label{sec:Arena-Lite}

To address the high annotation cost of Chatbot Arena while preserving evaluation reliability, we propose Arena-Lite. Arena-Lite introduces a tournament-based approach for efficient and reliable LLM evaluation using a single-elimination structure. Unlike baseline-mediated evaluations that compare model outputs to a baseline, Arena-Lite directly compares outputs from different models through head-to-head matchups for each prompt in benchmark datasets. Repeated tournaments across the dataset produce consistent leaderboards reflecting models' fundamental performance.

We first discuss limitations of baseline-mediated evaluations (Section \ref{sec:Comparing to Baseline outputs is not Always Helpful}). Next, we describe how Arena-Lite conducts tournaments to generate ratings (Section \ref{sec:Tournaments of LLMs over multiple prompts to preference ratings}, Algorithm \ref{alg:tournament}). Finally, we highlight similarities between the single-elimination structure and merge sort, explaining why aggregated tournaments yield reliable LLM rankings (Section \ref{sec:Why Aggregating Multiple Tournaments Yields Reliable Ranks}).

\subsection{Comparing to Baseline outputs is not Always Helpful} 
\label{sec:Comparing to Baseline outputs is not Always Helpful}

Although baseline outputs are a standard way to evaluate and rank LLMs, they introduce potential failure modes. Beyond the fact that a single baseline output might not capture every dimension of appropriate answers, relying solely on a baseline output can lead to unreliable rankings of LLMs.

Consider an ideal scenario with a judge capable of perfectly distinguishing the quality of any two outputs. If we choose to compare LLM responses directly to rank them using BT preference (Equation~\ref{eq:bt_model_elo}), all head-to-head comparisons are utilized. In contrast, baseline-mediated evaluation for differentiating LLMs can exhibit failure modes, as shown in Equation~\ref{eq:ref_useless_useful}.

\begin{equation}
\resizebox{\columnwidth}{!}{%
$\begin{array}{l@{\hspace{2em}}l}
\begin{array}{l}
\text{M}_1(X_i) \\
\quad \text{vs.}\, \rightarrow \\
\text{M}_2(X_i)
\end{array}
&
\begin{cases}
\text{M}_1(X_i) > Y_i > \text{M}_2(X_i) & \text{(helpful)} \\
\text{M}_1(X_i) < Y_i < \text{M}_2(X_i) & \text{(helpful)} \\
\text{M}_1(X_i),\ \text{M}_2(X_i) > Y_i & \text{(unhelpful)}\\
\text{M}_1(X_i),\ \text{M}_2(X_i) < Y_i & \text{(unhelpful)}
\end{cases}
\end{array}$%
}
\label{eq:ref_useless_useful}
\end{equation}

When the baseline output ($Y_i$) for a prompt ($X_i$) successfully disambiguates the pair of LLM responses $M_{1}(X_i)$ and $M_{2}(X_i)$ (as in the first and second cases), comparison to the baseline is effective for benchmarking. 
Otherwise, these comparisons do not help differentiate LLM performance. 
Consequently, the baseline-mediated approach provides less information for ranking when multiple responses are either both correct or both incorrect relative to the baseline.

\subsection{Tournaments of LLMs over multiple prompts to preference ratings}
\label{sec:Tournaments of LLMs over multiple prompts to preference ratings}

\begin{algorithm}
\caption{Tournament-Based Model Evaluation}
\label{alg:tournament}
\begin{algorithmic}[1]
\Require Models $M = \{m_1, m_2, \ldots, m_n\}$, Prompts $X = \{x_1, x_2, \ldots, x_k\}$
\Ensure Bradley-Terry preference ratings

\State Initialize $R \leftarrow \emptyset$

\For{each $x_j \in X$}
  \State $\texttt{next\_power} \leftarrow 2^{\lceil \log_2(|M|) \rceil}$
  \State $\texttt{n\_byes} \leftarrow \texttt{next\_power} - |M|$
  \State $M' \leftarrow M \cup \{\texttt{None}\}^{\texttt{n\_byes}}$
  \State Randomly shuffle $M'$
  \State $\texttt{winner} \leftarrow$ \textsc{SingleElim}($M'$, $x_j$)
\EndFor

\State \Return \textsc{ComputeBTM}($R$) \Comment{Eq. (1)}

\vspace{0.5em}
\Function{SingleElim}{$M'$, $x$}
  \If{$|M'| = 2$}
    \State $\texttt{result} \leftarrow$ \textsc{Match}($M'[0]$, $M'[1]$, $x$)
    \If{$\texttt{None} \notin M'$}
      \State Add $\texttt{result}$ to $R$
    \EndIf
    \If{$\texttt{result} = \texttt{None}$}
      \Return $[\texttt{None}]$
    \Else
      \Return $[\texttt{result}[0]]$
    \EndIf
  \Else
    \State $\texttt{mid} \leftarrow \lfloor |M'|/2 \rfloor$
    \State $\texttt{left} \leftarrow$ \textsc{SingleElim}($M'[:\texttt{mid}]$, $x$)
    \State $\texttt{right} \leftarrow$ \textsc{SingleElim}($M'[\texttt{mid}:]$, $x$)
    \Return \textsc{SingleElim}($\texttt{left} \cup \texttt{right}$, $x$)
  \EndIf
\EndFunction

\vspace{0.5em}
\Function{Match}{$m_i$, $m_j$, $x$}
  \If{$m_i = \texttt{None}$ and $m_j = \texttt{None}$}
    \State \Return $\texttt{None}$
  \ElsIf{$m_i = \texttt{None}$ or $m_j = \texttt{None}$}
    \State \Return $m_i$ if $m_j = \texttt{None}$ else $m_j$
  \Else
    \State $O_{i} \leftarrow m_i(x)$, $O_{j} \leftarrow m_j(x)$
    \If{$O_{j} > O_{i}$}
      \Return $(m_j, m_i)$
    \Else
      \Return $(m_i, m_j)$\\ \Comment{returns (winner, loser) tuple}
    \EndIf
  \EndIf
\EndFunction

\end{algorithmic}
\end{algorithm}

Figure~\ref{fig:va_conceptual} and Algorithm~\ref{alg:tournament} illustrate how Arena-Lite benchmarks LLMs via a tournament approach. Here, $|X|$ denotes the number of prompts in the benchmark dataset. Running Arena-Lite hosts tournaments among participant LLMs for every prompt in the dataset.

The use of tournament structures for LLM benchmarking offers both benefits and challenges. 
A major advantage of a single-elimination tournament is efficiency. As shown in Table~\ref{table:comparison}, the number of matches scales linearly with the number of participants and even lower compared to using baseline outputs. 
However, single elimination tournament only identifies a champion, leaving the relative ordering of other participants unclear.

To retain tournament's efficiency while obtaining a fine-grained ranking, we propose aggregating tournament results over multiple prompts with randomized initial match-ups for each prompt. Performing multiple tournaments with random initialization offers several benefits: 
\begin{enumerate} 
\item It resolves ties among non-champion participants from previous tournaments. 
\item It mitigates the impact of unfavorable match-ups in any single tournament. 
\item Aggregating match results allows for precise win rate estimation via BT preference, resulting in a well-aligned overall ranking. 
\item More matches are allocated to high-performing participants while ensuring every participant is evaluated at least once per prompt (Table~\ref{table:comparison}). 
\end{enumerate}

\noindent In Section~\ref{sec:Why Aggregating Multiple Tournaments Yields Reliable Ranks}, we further explain how aggregating multiple tournaments could yield reliable ranking of LLMs. 
We also provide further analysis on number of comparisons performed over tournaments of Arena-Lite comparing to merge sort, offering a comprehensive view of the method's efficiency and effectiveness.

\subsection{Why Aggregating Multiple Tournaments Yields Reliable Rankings}
\label{sec:Why Aggregating Multiple Tournaments Yields Reliable Ranks}


A key challenge in LLM evaluation is to derive a reliable ranking from a feasible number of pairwise comparisons. 
Our tournament-based approach, Arena-Lite, is designed to efficiently sample these comparisons. 
In a single tournament with n models, each model participates in a minimum of one match and a maximum of $\lceil \log_2 n_\text{model} \rceil$ matches. 
When conducted over a benchmark with $|X|$ distinct prompts, the total number of evaluations per model is bounded within the interval $[|X|,~|X|* \lceil \log_2 n_\text{model} \rceil ]$ as presented in Table~\ref{table:comparison}. 
However, the efficiency of this tournament structure raises a critical question: how does this limited sampling produce a statistically reliable global ranking? 

To achieve reliable rankings of LLMs, our approach aggregates match outcomes from multiple tournaments, lesser than a full grid comparisons, but effectively approximating the complete set of pairwise comparisons required for merge sort. We outline the rationale in four key points:
\newline

\noindent \textbf{Merge Sort Baseline} A single-elimination tournament mirrors the merging steps of merge sort, which requires \( \mathcal{O}(n \log n) \) comparisons with no duplicate match-ups to rank \( n \) models. However, a single tournament omits many comparisons, covering only the minimal match-ups needed to determine a winner.

\noindent \textbf{Recovering Comparisons via Aggregation} By aggregating tournaments over diverse prompts, we can recover missed pairwise match-ups that should have occurred.
Assuming match outcomes are prompt-independent (as BT model assumes), matches across prompts are considered equivalent. With \( |X| \) prompts (typically hundreds to thousands) and \( n_{\text{model}} \) models (tens), only random initial match-ups totaling \( |X| \cdot \lfloor \frac{n_{\text{model}}}{2} \rfloor \). This exceeds the \( \binom{n_{\text{model}}}{2} \) possible combination, ensuring broad coverage.

\noindent \textbf{Sufficiency of Comparisons} The aggregated match-ups not only cover the necessary comparisons but also surpass the \( \mathcal{O}(n \log n) \) requirement of merge sort. Moreover, each unique model pair competes in average $|X|/(n_\text{model}-1)$ to $\frac{|X|\log_2 n_\text{model}}{n_\text{model}-1}$ matches\footnote{These estimates are computed by number of matches a model undergoes which is $[~|X|,~|X|* \lceil \log_2 n_\text{model} \rceil ~]$, devided by number of possible opponents for a model, $n_\text{model}-1$} across the benchmark, a frequency mostly sufficient for accurate win rate estimation.

\noindent \textbf{Refinement for Reliability} The remaining matches, totaling \( |X| \cdot (n_{\text{model}}-1) \), further refine the ranking by enhancing win rate estimates, especially among top-performing models, reducing noise and ensuring robustness akin to Arena-Lite’s sampling strategy.

In summary, aggregating multiple tournaments reconstructs the full set of comparisons needed for a merge sort-like ranking while providing enough repeated match-ups to ensure accurate win rate estimations. This dual mechanism yields reliable and robust LLM rankings across the benchmark.

\section{Experiments}
\label{sec:experiments}

We conducted two experiments to evaluate Arena-Lite against baseline-mediated benchmarking. The first experiment (Section \ref{sec:exp1}) utilized a stochastic model to simulate LLM competitions, comparing Arena-Lite's tournament-based direct comparison with baseline-mediated evaluation. This controlled setup allowed us to test Arena-Lite's design principles, such as the effectiveness of direct versus mediated comparison (Section \ref{sec:Comparing to Baseline outputs is not Always Helpful}) and tournament-based sampling (\ref{sec:Why Aggregating Multiple Tournaments Yields Reliable Ranks}), while isolating variables and minimizing noise, such as LLM judge biases \cite{park2024offsetbiasleveragingdebiaseddata}.
The second experiment (Section \ref{sec:exp2}) validates Arena-Lite empirically using various LLMs as judges and public benchmark data. We tested models including \texttt{gpt-4o}, \texttt{gpt-4o-mini}, \texttt{Claude3.5}, \texttt{Qwen2.5}, \texttt{Llama3.1}, and \texttt{Gemma2} to assess Arena-Lite’s effectiveness against standard benchmarking practices. Together, these experiments demonstrate the superior reliability and efficiency of Arena-Lite’s tournament approach.
Section \ref{sec:Ratings from Chatbot Arena Leaderboard as a Ground-Truth LLM Rankings} outlines shared experimental settings, followed by detailed descriptions of each experiment in subsequent subsections.
\subsection{Chatbot Arena Leaderboard as Ground-Truth Rankings}
\label{sec:Ratings from Chatbot Arena Leaderboard as a Ground-Truth LLM Rankings}
We benchmark Arena-Lite and baseline-mediated evaluation against rankings from the Chatbot Arena leaderboard, widely recognized for its reliability due to extensive human preference annotations. With a large volume of votes across diverse prompts, these rankings provide a robust ground truth for model comparisons.

\subsection{Experiment 1: Controlled Stochastic Modeling of LLM Competitions}
\label{sec:exp1}

We suggest a simple stochastic model based on the Bradley-Terry (BT) framework to compare Arena-Lite’s approach with baseline-mediated evaluation. This experiments simulates prompt-agnostic LLM competitions which follows the Bradley-Terry model's presumption, with outcomes determined by a judge following Equation \ref{eq:correct_outcome_likelihood}. The judge’s decision is based on the BT preference difference ($\Delta_\text{ij}$) between models $i$ and $j$, and the judge’s accuracy ($P_\text{judge}$):

\begin{equation}
\begin{split}
P_\text{predict}(i > j) &= P_\text{judge} \times P_\text{gt}(i > j) \\
            &= P_\text{judge} \times \frac{1}{1 + 10^{\Delta_\text{ij}/400}}
\end{split}
\label{eq:correct_outcome_likelihood}
\end{equation}

\noindent With the model of judge above (Equation \ref{eq:correct_outcome_likelihood}), we simulate both Arena-Lite's tournament-based approach and baseline-mediated approaches according to the following initial conditions and procedures. 
\newline
\linebreak
\noindent \textbf{Initial conditions:}

\begin{itemize}
  \item \textbf{Ground-Truth BT Preference}: We extracted BT preferences from the English category of Chatbot Arena (as of June 23), derived from approximately 60\% of user-submitted judgments. These preferences serve as both the initial model parameters and the ground-truth rankings for evaluation.
  \item \textbf{Judge Accuracy ($P_\text{judge}$)}: We varied judge accuracy from 0.6 to 0.9.
  \item \textbf{Number of LLMs ($n_\text{model}$) and Dataset Size ($|X|$)}: We adjusted the number of participating LLMs and benchmark dataset sizes to assess the robustness of both approaches in data-poor and data-rich settings.
\end{itemize}

\noindent \textbf{Simulation Procedure:}

\begin{enumerate}
  \item Select participant LLMs and their BT preferences.
  \item Compute expected win rates ($P_\text{gt}$) using Equation \ref{eq:correct_outcome_likelihood}.
  \item Sample match outcomes based on $P_\text{predict}$ (Equation \ref{eq:correct_outcome_likelihood}), determined by the BT preference rating gap ($\Delta_\text{ij}$) and judge accuracy ($P_\text{judge}$).
  \item Repeat for the specified number of test prompts ($|X|$).
  \item Compute scores:
    \begin{itemize}
      \item \textbf{Baseline-mediated}: Win rate against a reference model (\texttt{gpt-4-1106-preview}, rating 1233).
      \item \textbf{Arena-Lite}: BT preference from all tournament match outcomes.
    \end{itemize}
  \item Rank models based on scores.
  \item Calculate Spearman correlation between simulated and ground-truth rankings.
\end{enumerate}

We conducted 50 trials per configuration to account for stochasticity in initial tournament brackets and judging process.

\subsection{Experiment 2: Empirical Validation of Arena-Lite with real LLM Judge}
\label{sec:exp2}
To empirically validate our proposal, we evaluated the reliability of both Arena-Lite and baseline-mediated approach over the top 19 models from the Chatbot Arena leaderboards. This experiment employs actual prompt inputs and LLM outputs, distinguishing it from the earlier simulation study.

\subsubsection{Dataset: Test Prompts and LLM Responses Used}
Testing the benchmarking approaches requires: (1) test prompts and (2) the corresponding responses from LLMs. For the benchmark dataset, we selected Arena-Hard-Auto~\cite{arena-hard}. 
The prompts in Arena-Hard-Auto were carefully curated from Chatbot Arena user queries. 
This dataset consists of 500 prompts—two instances for each of 250 subtopics. 
Although AlpacaEval~\cite{alpacaeval}, which comprises 800 prompt-reference pairs, could serve as a viable testbed, we opted for Arena-Hard-Auto because its design aligns more closely with Chatbot Arena.
Arena-Hard-Auto uses responses from \texttt{gpt-4-0314} as the baseline outputs. 
For ranking, we utilized the reserved outputs of the top 20 (=19 + baseline) models from the Arena-Hard-Auto Browser.\footnote{Extracted from the 2024 Jul 6 commit (\texttt{fd42026}).}

\subsubsection{Participant LLMs}
For ranking, we selected 19 LLMs from the top of the ChatBot Arena leaderboard in the \textit{hard prompts} category, as these models most closely align with Arena-Hard-Auto. 

\subsubsection{LLM Judges} 
We used several aligned LLMs as judges for testing both benchmarking approaches.
LLMs of our choice are \texttt{gpt-4o} family of models~\cite{openai2024gpt4ocard}, \texttt{Claude3.5}, and a selection of open-weight models: \texttt{Qwen2.5}~\cite{qwen2025qwen25technicalreport}, \texttt{Llama3.1}~\cite{grattafiori2024llama3herdmodels}, and \texttt{Gemma2}~\cite{gemmateam2024gemma2improvingopen}. 
For pairwise comparisons of responses, we employed the judging prompt suggested in LLMBar~\cite{LLMBar} (See Appendix \ref{fig:prompt}). 
The same judge prompt was applied consistently across both the tournament and baseline-mediated approaches. 
To mitigate position bias~\cite{posbias}, the order of model responses was alternated during evaluation. 
Further details on the LLM-as-a-judge configuration are provided in Appendix~\ref{appendix:Prompts and Judge configuration}.
\\
\\
The two experimental settings are summarized as follows:
\newline

\noindent \textbf{Experiment 1 (Modeling Experiment):}
This experiment uses the ground truth BT preference of the models to initialize the simulation. 
We vary control parameters for the benchmarking approaches—including the judge's accuracy ($P_\text{judge}$), the number of test prompts used ($|X|$), and the number of participant LLMs ($n_\text{model}$)—to determine which benchmarking approach more accurately reproduces the participants' ranking.
For each configuration, we conduct 50 trials of experiments.

\noindent \textbf{Experiment 2 (Empirical Validation):}
This experiment assesses the two benchmarking approaches using empirical runs with various LLM judges. We select the top 19 LLMs from Chatbot Arena and used their reserved outputs on Arena-Hard-Auto test prompts. For both the tournament and baseline-mediated approaches, we employ the Spearman correlation coefficient to measure how well the results align with the ground truth leaderboard rankings. In our empirical study, we conduct 500 trials for each experimental setting. 

\begin{figure*}
  \includegraphics[width=\textwidth]{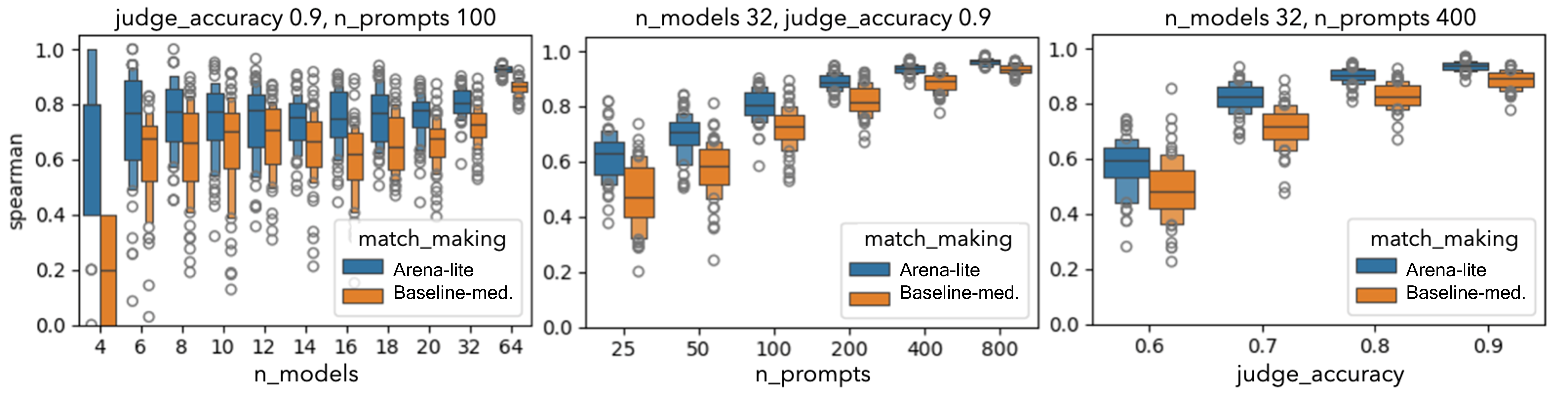}
 \caption{
Comparison of LLM ranking reliability between Arena-Lite and a baseline method in a stochastic simulation (Experiment 1, Sec. \ref{sec:exp1}). Ranking reliability is measured by the Spearman correlation ($\uparrow$) between the competition-derived ranking and the ground-truth ranking. Each box plot summarizes the results from 50 trials. The subplots analyze the effect of varying (from left to right) the number of competing models ($n_\text{models}$), the number of prompts ($n_\text{prompts}$), and the accuracy of the judge ($P_\text{judge}$). The single-elimination structure of Arena-Lite results in consistently higher correlation scores.
 }
  \label{fig:exp1fig}  
\end{figure*}

\section{Results and Discussion}
\label{sec:results}
We assess the reliability and robustness of Arena-Lite as a means for LLM benchmarking, comparing it against the current baseline-mediated approach. 
Our results from both simulation study and empirical runs indicate that the tournament approach of Arena-Lite yields rankings that align more closely with the ground-truth Chatbot Arena leaderboards. 
We present our findings using whisker plots and tables in the following sections.

\subsection{Experiment 1: Modeling Experiment Results}

Figure \ref{fig:exp1fig} illustrates noticeable differences in Spearman correlation, indicating that the tournament approach is more reliable than the baseline-mediated method. 
The consistent performance gap across various conditions—namely, the number of participants, the number of test prompts, and judge accuracy ($n_\text{model}$, $|X|$, and $P_\text{judge}$)—demonstrates the robustness of the tournament approach. 
Although the simulation simplifies real-world complexity, a similar performance gap was observed in the empirical findings (Experiment 2, Figure \ref{fig:exp2fig}). 
This consistency suggests that the robust performance of Arena-Lite is not coincidental or limited to a specific empirical setting of ours.

\subsection{Experiment 2: Empirical Validation Results}
\begin{figure}
  \centering
  \includegraphics[width=1\linewidth]{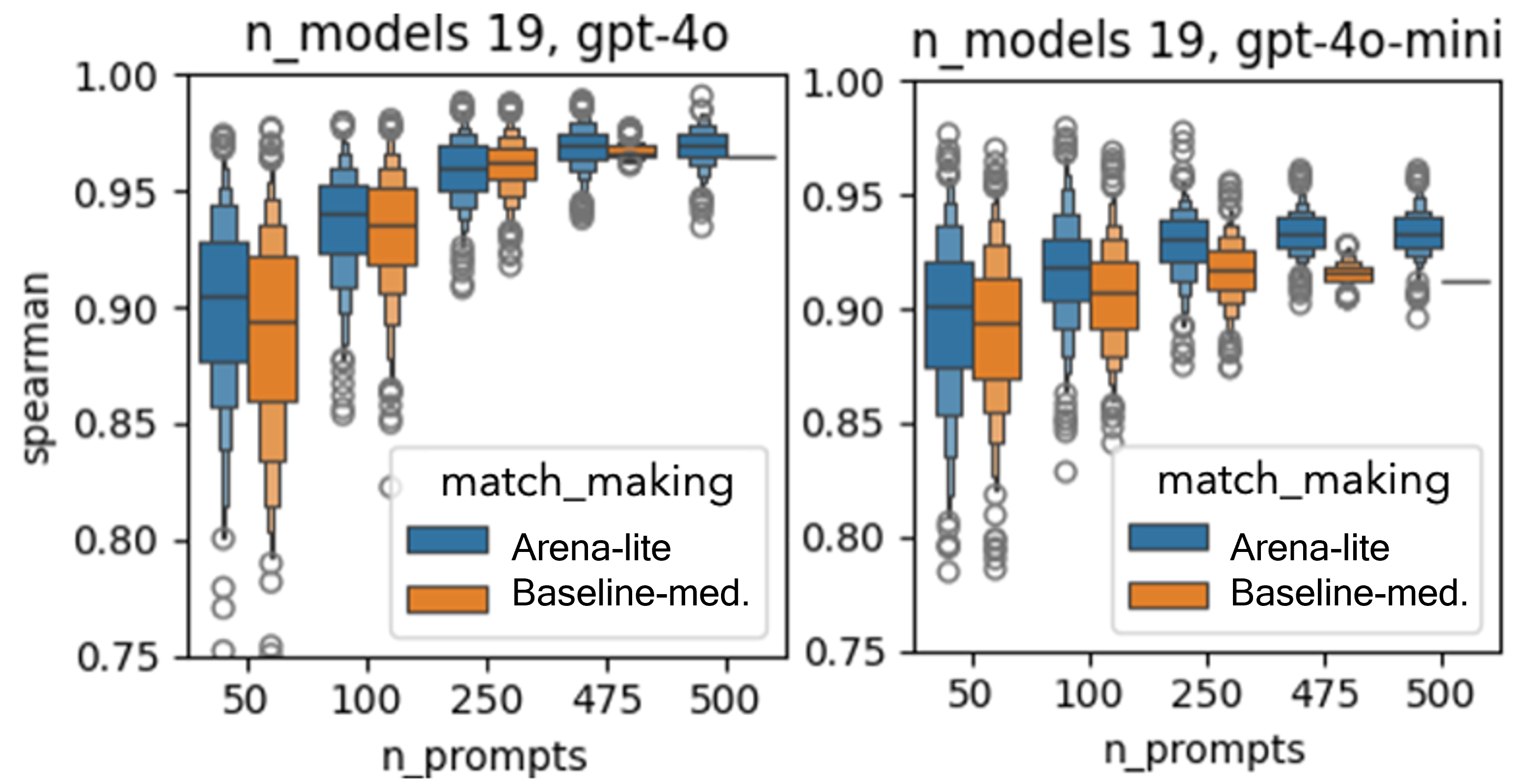}
  \caption{Ranking reliability of Arena-Lite vs. utilizing baseline outputs. Arena-Lite consistently demonstrates higher Spearman's rank correlation across numbers of benchmark prompts ($|X|$), indicating more reliable ranking. The evaluation was performed using \texttt{gpt-4o} (left) and \texttt{gpt-4o-mini} (right) as judge models, with a fixed number of models ($n_\text{models}$=19). Each box plot summarizes the results of 50 runs. (Experiment 2, Sec. 4.3).}
  \label{fig:exp2fig}
\end{figure}

As hinted in the previous section, the empirical results in Figure~\ref{fig:exp2fig} show that Arena-Lite consistently outperforms the baseline-mediated approach. 
Although the performance gaps are less pronounced than in the simulation, the same trend persists. 
In Table~\ref{table:anchored_vs_tournament_n_prompts}, we report the median values for Arena-Lite and the baseline-mediated approach using the \texttt{gpt-4o} family of judges while varying the number of test prompts ($|X|$). 
These results consistently demonstrate that Arena-Lite outperforms the baseline-mediated method.
Note that Arena-Lite shows similar or superior reliability even in extreme data-poor benchmark condition ($|X|=50$).

Table~\ref{tab:other LLM result} presents the outcomes when using other LLMs as judges, with a fixed number of prompts ($|X|=500$). 
The results for \texttt{Claude3.5-sonnet}, \texttt{Llama3.1-8b}, and \texttt{Qwen2.5-7b} follow a similar trend. 
However, smaller models (\texttt{Gemma2-2b} and \texttt{Qwen2.5-0.5b}) appears to be less reliable as an LLM judge. 
Hence, we recommend using evaluation-specialized judge LLMs or, at least, generative judge models with around 7B parameters regardless of using Arena-Lite or considering baseline-mediated approach.

\begin{table}[ht]
\centering
  \resizebox{\linewidth}{!}{
  \begin{tabular}{lccccc}
  \hline
  Spearman corr. ($\uparrow$) & $|X|$ = 50& 100 & 250 & 475 & 500 \\
  \hline
  baseline-mediated (4o) & 0.895 & 0.935 & \textbf{0.963} & 0.966 & 0.964 \\
  Arena-Lite (4o) & \textbf{0.905} & \textbf{0.940} & 0.960 & \textbf{0.970} & \textbf{0.970} \\
  \hline
  baseline-mediated (4o-mini) & 0.895 & 0.908 & 0.917 & 0.916 & 0.912 \\
  Arena-Lite (4o-mini) & \textbf{0.901} & \textbf{0.919} & \textbf{0.931} & \textbf{0.933} & \textbf{0.933} \\
  \hline
  \end{tabular}
  }
  \caption{Robustness of ranking methods to benchmark set size, $|X|$ (Experiment 2, Sec. \ref{sec:exp2}). The table shows the median Spearman correlation ($\uparrow$) from 500 trials. Arena-Lite consistently achieves higher correlation than the baseline-mediated approach across all dataset sizes ($|X|$), demonstrating its superior reliability and robustness for ranking LLMs. 
  }
\label{table:anchored_vs_tournament_n_prompts}
\end{table}

\begin{table}[ht]
\centering
\resizebox{\linewidth}{!}{
\begin{tabular}{lccccc}
\hline
 $|X|$ = 500& \makecell{claude3.5\\sonnet}& \makecell{llama3.1\\8b-it} & \makecell{qwen2.5\\7b-it} & \makecell{qwen2.5\\0.5b-it} & \makecell{gemma2\\2b-it} \\
\hline
baseline-mediated & 0.924 & 0.820& 0.756 & 0.089& \textbf{0.592}\\
Arena-Lite  & \textbf{0.930}& \textbf{0.850}& \textbf{0.811}& -0.124 & 0.552 \\
\hline
\end{tabular}
}
\caption{
Robustness of ranking methods to the choice of judge LLM (Experiment 2, Sec. \ref{sec:exp2}). The table shows the Spearman correlation ($\uparrow$) between ground-truth LLM rankings and the results from each method. The values are medians from 500 trials. The results suggest that around 7B parameters-large LLMs is a viable minimum threshold for a reliable judge. Full results for other benchmark sizes are available in Appendix, Table~\ref{tab:appendix_eval_results}.}
\label{tab:other LLM result}
\end{table}

\subsection{Incorporating a New LLM into an Existing Leaderboard}

While our main focus has been on ranking multiple LLMs at once, it is also useful to consider the common scenario of adding a single new model to an existing leaderboard, which is also frequent use-case. 
We explored two approaches: (1) a \textit{binary search}-like placement method, and (2) using the top-performing model response as a baseline. 
Our findings indicate that the later approach is more reliable (Table~\ref{tab:binarysearch_vs_anchored}, Appendix). 
Further details and discussions are provided in Appendix~\ref{appendix:Binary search vs. Win rate over baseline}.




\section{Related Works}
\label{sec:related}
\subsection{LLM-as-a-Judge for Systems Ranking}
Utilizing LLM-as-a-Judge as a building block for systems ranking has become a common practice in the LLM benchmarking community. Several studies have investigated how LLM judges compare to human evaluators, examining their similarities and differences~\cite{park2024offsetbiasleveragingdebiaseddata}, as well as how these differences impact system rankings (e.g., JuStRank~\cite{gera2024justrank},~\cite{gao-etal-2025-evaluating}). Our research extends these approaches by proposing a method that orchestrates LLM-as-a-Judge through a well-established tournament structure to derive rankings among systems.

\subsection{Efficient and Reliable Evaluation}
There is a growing body of research focused on optimizing the number of evaluations while maintaining reliability when using LLM-as-a-Judge for system ranking. \citeauthor{perlitz-etal-2024-efficient} proposed a metric called DIoR to quantify the relationship between computational costs and system ranking reliability. UniCBE~\cite{yuan2025unicbeuniformitydrivencomparingbased} introduced a method to analyze the relationship between reliability and the number of judge evaluations based on uncertainty. BenchBench~\cite{perlitz2024llmbenchmarksagreefixing} systematically analyzed consistency across benchmarks and provided a package to facilitate this analysis. tinyBenchmarks~\cite{polo2024tinybenchmarks} explored strategies to minimize the number of evaluations across various established benchmarks. Arena-Lite relates to these studies in that it leverages the properties of tournament structures and direct comparisons to achieve more reliable results with fewer judge evaluations.

\section{Conclusion}
\label{sec:conclusion}
We introduced Arena-Lite, an efficient and reliable framework for evaluating Large Language Models (LLMs) through tournament-based direct comparisons. By eliminating the need for baseline outputs and adopting head-to-head comparison, Arena-Lite achieves higher reliability in system rankings with reduced number of comparisons. Our experiments, encompassing controlled stochastic modeling and empirical validation with various LLM judges, confirm that Arena-Lite consistently outperforms standard baseline-mediated evaluation methods, even with smaller datasets or weaker judges. The release of an accessible web demonstration and code supports the adoption of Arena-Lite to help streamlining model development cycle across research and industry. Future work will extend Arena-Lite's application to diverse domains, including multi-modal LLM evaluation involving visual or audio inputs and outputs.

\section*{Limitations}
While we conducted extensive testing to assess the robustness of Arena-Lite tournaments—including 50 and 500 trials for Experiment 1 and Experiment 2, respectively—some inherent sources of randomness remain, such as variation due to initial match bracket assignments. The randomness in bracket assignment is added for adopting tournament structure of Arena-Lite and may influence outcome stability. Future work could explore more informative or adaptive matchmaking strategies that improve ranking fidelity beyond what is achievable with single-elimination formats, potentially within the same or even fewer number of matches.

\bibliography{custom}

\appendix

\section{Appendix}

\subsection{Arena-Lite Web Demo}
We provide screenshots of Arena-Lite web demo here (\href{https://huggingface.co/spaces/NCSOFT/ArenaLite}{link to Huggingface Space}). You could try or locally host Arena-Lite according to its documentation. It is quite easy to use and do not require much of resources to host. 

\begin{figure}[h]
  \centering
  \includegraphics[width=1\linewidth]{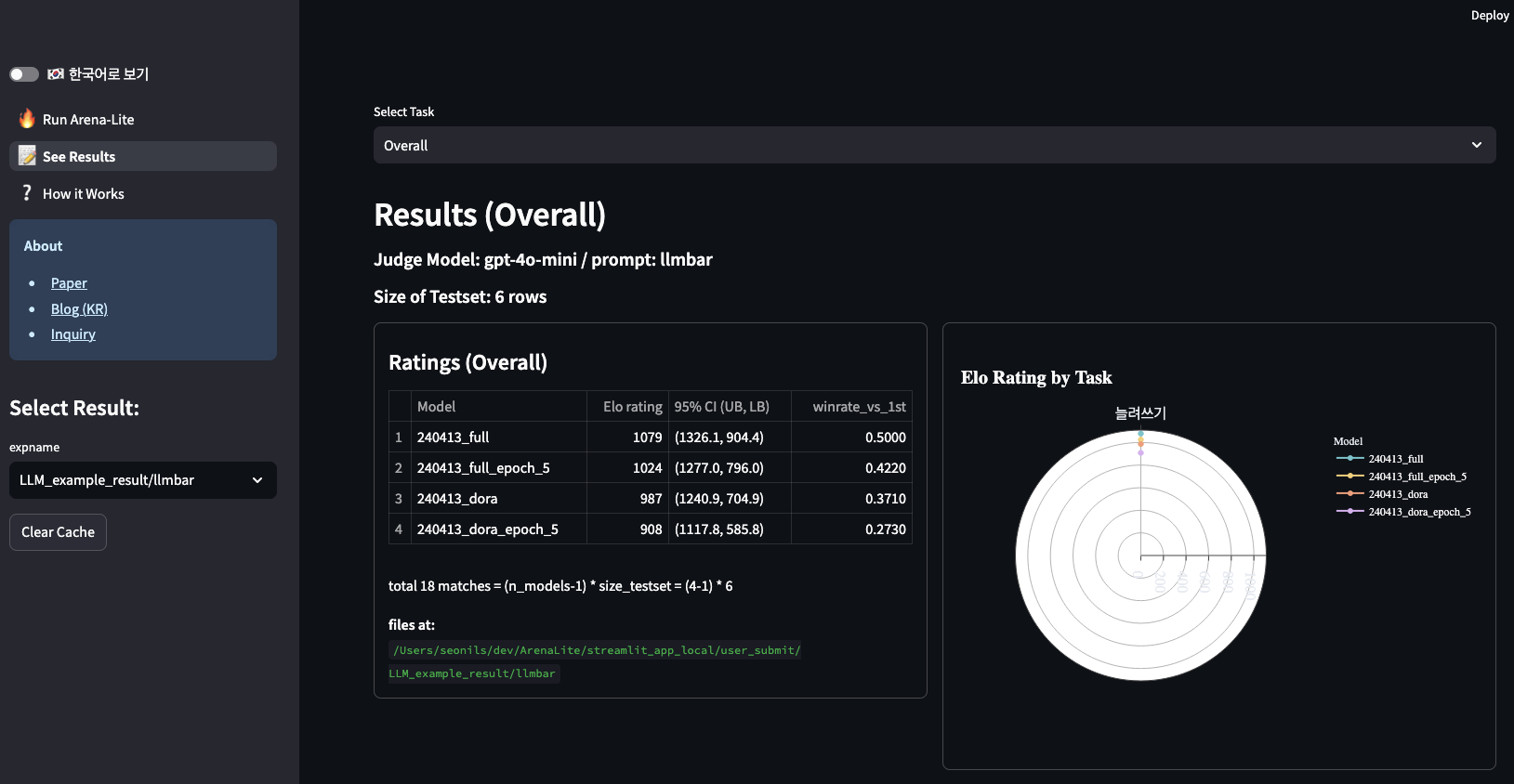}
  \caption{Arena-Lite web screenshot 1: At the top of the result page, one can see the leaderboard of LLMs with their BT preference. If the benchmark dataset has subcategories, radar chart (right) is also visible.}
  \label{fig:demo1}
\end{figure}

\begin{figure}[h]
  \centering
  \includegraphics[width=1\linewidth]{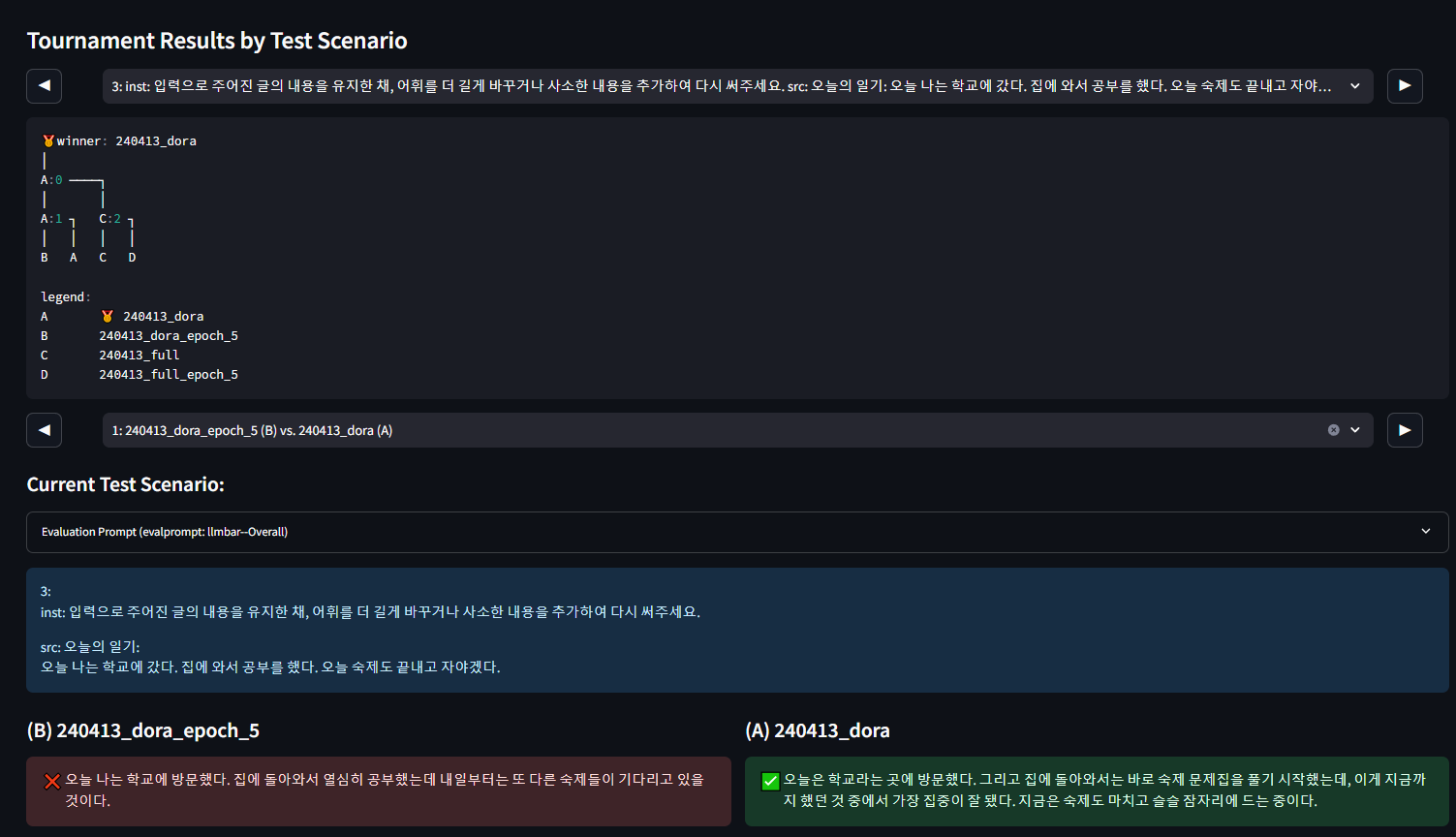}
  \caption{Arena-Lite web screenshot 2: User can walk through the matches and tournaments one by one. Match brackets is visualized briefly with text UI and user can select any specific match to see the details (e.g. match result, prompt, and model outputs).}
  \label{fig:demo2}
\end{figure}

\begin{figure}[h]
  \centering
  \includegraphics[width=1\linewidth]{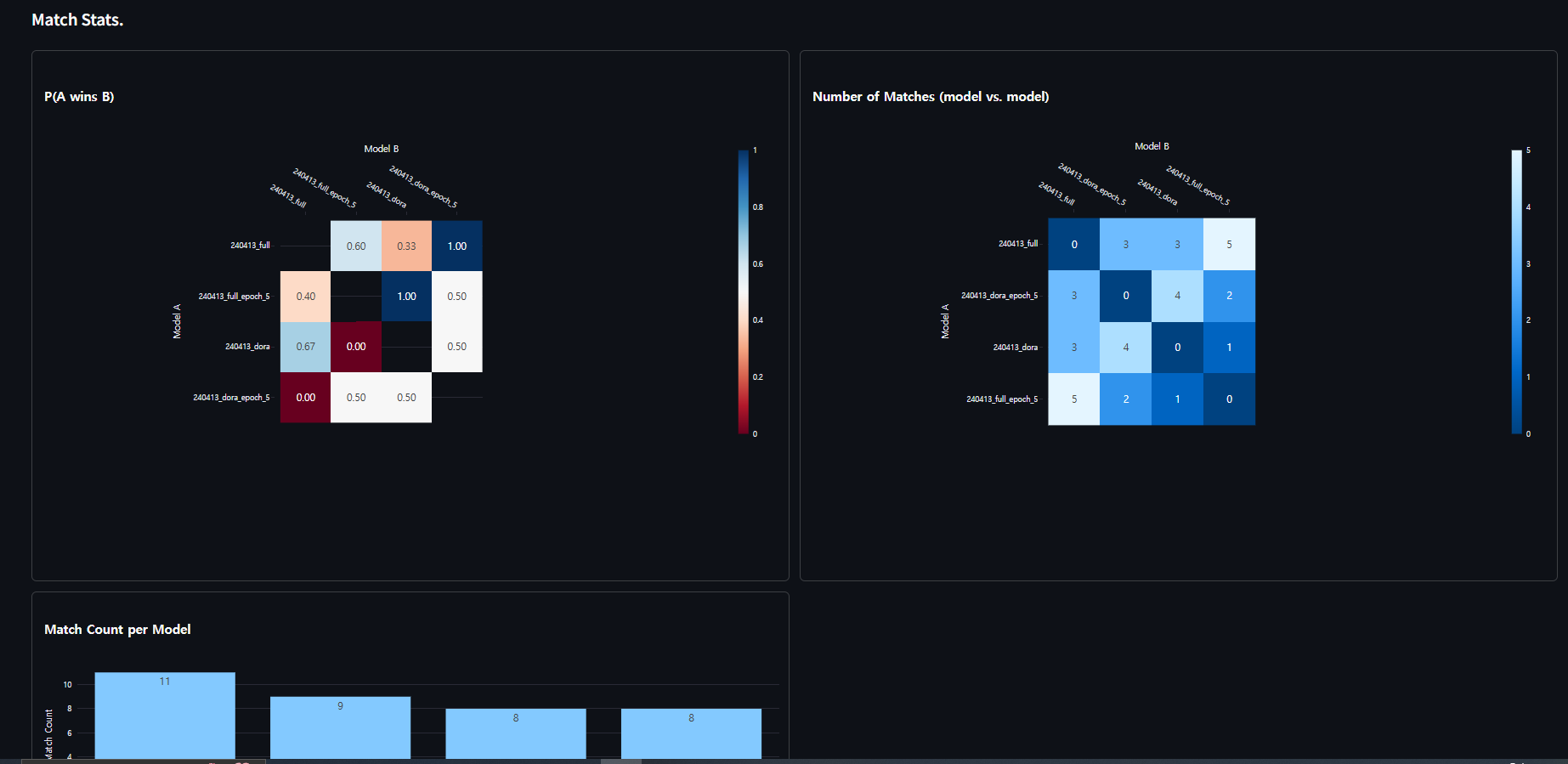}
  \caption{Arena-Lite web screenshot 3: User can see the match statistics between LLMs (i.e. win rate between model pairs, number of matches per pair and per model).}
  \label{fig:demo3}
\end{figure}

\begin{figure}[h]
  \centering
  \includegraphics[width=1\linewidth]{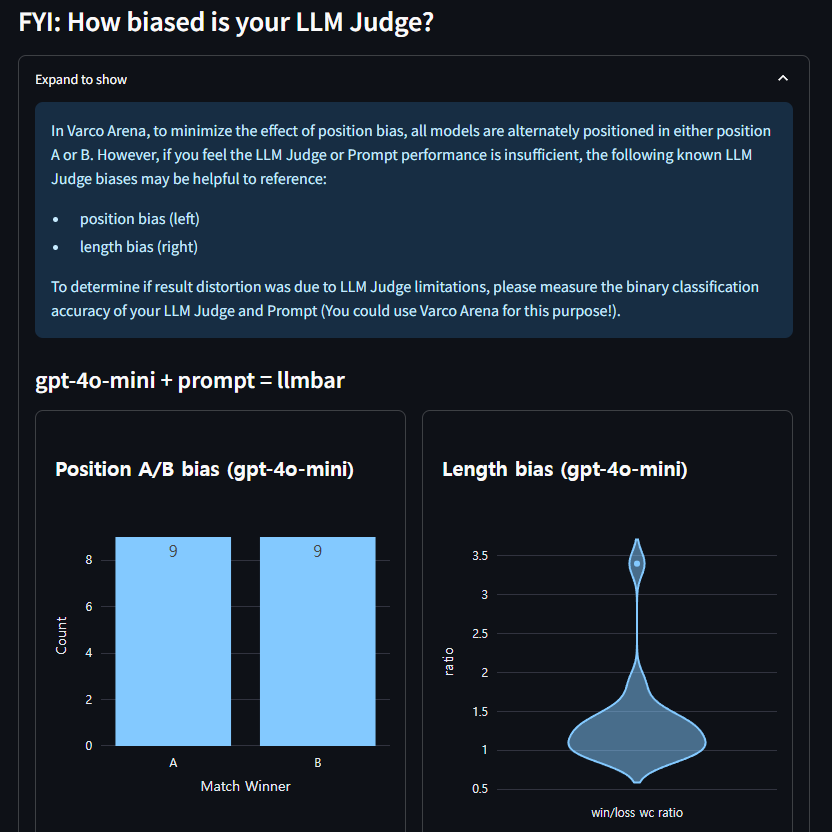}
  \caption{Arena-Lite web screenshot 4: User can see the LLM Judge's examine how biased the LLM judge being used. The demo provides clues for potential bias toward response length and position.}
  \label{fig:demo4}
\end{figure}

Arena-Lite provides the benchmark result (Figure~\ref{fig:demo1}) with helpful visualization interface that enables walking through the matches and tournaments one by one (Figure~\ref{fig:demo2}) and match statistics between LLMs (Figure~\ref{fig:demo3}). We also provide visualization that helps examining potential bias of LLM Judge being used (Figure~\ref{fig:demo4}).

\subsubsection{Starter Prompt Set}
We provide several judge prompts that we have used for specific target tasks. Some of those are for quite specialized tasks, and some might work for evaluating general instruction following. You could customize your own judge prompt based on your evaluation needs according to the documentation. The list of the judge prompts we provide is as follows, and one could see the detailed prompts at yaml files \href{https://huggingface.co/spaces/NCSOFT/ArenaLite/tree/main/varco_arena/varco_arena_core/prompts}{here}. 
\begin{enumerate}
\item  \texttt{llmbar} prompt (Figure~\ref{fig:prompt}) and conciser version of the prompt, \texttt{llmbar\_brief}. Those are for evaluating instruction following.
\item  \texttt{translation\_pair} prompt for selecting translation models trained on game-specialized parallel corpora,
\item  \texttt{rag\_pair\_kr} prompt for evaluating knowledge groundedness of korean RAG models over chatting scenario, 
\item  \texttt{translation\_fortunecookie} prompt which was crafted for evaluating translation models specialized for translating fortune-tellings, and 
\item  \texttt{post\_edit} prompt for evaluating conversation revision based on given persona of a speakers given in a instruction, which is quite specialized use case. 
\end{enumerate}

\subsection{Full table for Experiment 2}
Here is the extended results of Experiment 2 (Section \ref{sec:exp2}) presented in Table \ref{tab:other LLM result}. Aligned LLMs smaller than 7B parameters struggles to work as a proper Judge. Otherwise, Arena-Lite method excels over common practice of using outputs from proprietary as baselines.

\clearpage

\begin{table*}[htbp]
\centering
\footnotesize
\begin{tabular*}{\textwidth}{@{\extracolsep{\fill}}clccccc@{}}
\toprule
\textbf{Dataset size} & \textbf{method} & \textbf{claude 3.5} & \textbf{llama3.1} & \textbf{qwen2.5} & \textbf{qwen2.5} & \textbf{gemma2} \\
& & \textbf{sonnet} & \textbf{8b-it} & \textbf{7b-it} & \textbf{0.5b-it} & \textbf{2b-it} \\
\midrule
50 & baseline-mediated   & .896 & .656 & .492 & .010 & .064 \\
  & Arena-Lite (ours) &\textbf{ .897} & \textbf{.715} & \textbf{.544} & -0.051 & -0.088 \\ \midrule
100 & baseline-mediated   & .912 & .732 & .596 & .002 & .079 \\
  & Arena-Lite (ours) & \textbf{.918} & \textbf{.780} & \textbf{.656} & -0.068 & -0.090 \\ \midrule
250 & baseline-mediated   & .924 & .801 & .700 & .045 & \textbf{.560} \\
  & Arena-Lite (ours) & \textbf{.929} & \textbf{.830} & \textbf{.760} & -0.131 & .551 \\ \midrule
475 & baseline-mediated   & .924 & .819 & .708 & .083 & .112 \\
  & Arena-Lite (ours) & \textbf{.930} & \textbf{.845} & \textbf{.810} & -0.131 & -0.009 \\ \midrule
500 & baseline-mediated   & .924 & .820 & .756 & .089 & \textbf{.592} \\
  & Arena-Lite (ours) & \textbf{.930} & \textbf{.850} & \textbf{.811} & -0.124 & .551 \\
\bottomrule
\end{tabular*}
\caption{Extended results for comparing Arena-Lite to baseline-mediated method of using outputs from proprietary models as an baseline. We tested other LLMs as judge over various size of benchmark datasets.}
\label{tab:appendix_eval_results} 
\end{table*}

\clearpage

\subsection{Machine Requirements for Experiments}
Except the part we inferenced open-weight models such as \texttt{Llama, Qwen} and \texttt{Gemma}, our experiments are mostly do not require GPU usage.
Inference are done on one A100 GPU, but T4 would be enough for reproducing our experiments.
Otherwise, our experiments require querying API and post-processing those with CPU. Experiments could be run on personal desktops. The lowest specification of the machine we deployed had \texttt{i5-8400 CPU}, 16 GiB RAM.


\subsection{Assuring Statistical Significance of the Results within Budget for proprietary models}
To ensure a statistically significant number of trials for each experiment while staying within budget, we utilize OpenAI's Batch API to prepare full-grid match outcomes (i.e., all-play-all matches for every prompt) in a cache file, allowing us to reuse these outcomes. 
Each empirical experiment consists of 500 trials per setting, with results represented using whisker plots or summary statistics such as median values. 
When experimenting with a subset of the Arena-Hard-Auto benchmark ($|X|<500$), we sample a stratified subset of the benchmark dataset for each new trial. 

\begin{figure*}
  \centering
  \includegraphics[width=1\textwidth]{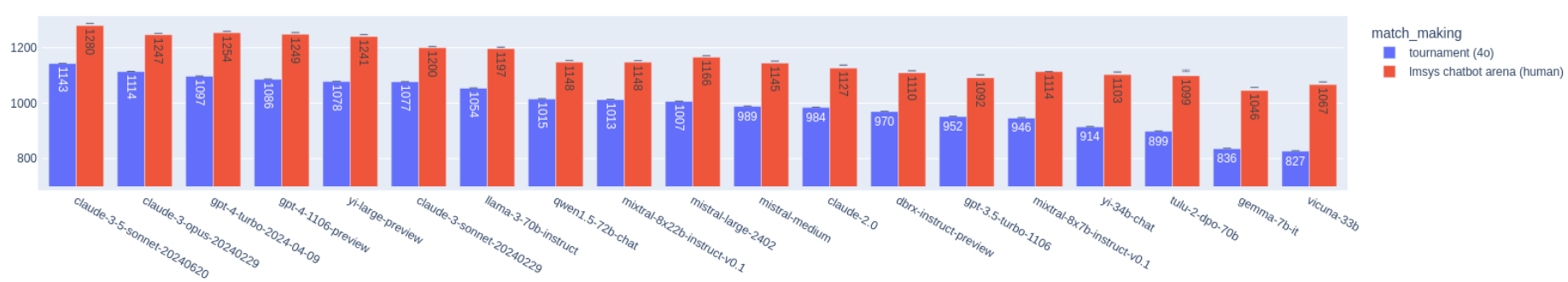}
  \caption{BT preference of the model with gpt-4o judge on the full set of Arena-Hard-Auto~\cite{arena-hard} prompts. Arena-Lite result (bootstrapped median over 1000 samples of 500 trials) is in blue, plotted alongside the ratings from the ground truth leaderboard in red (Chatbot Arena, \textit{Hard prompts category}). Error bars are 95\% confidence intervals.}
  \label{fig:bar_with_ci_vs_chatbot_arena_gpt4o}
\end{figure*}

\subsection{BT preference from Arena-Lite compared to Human Annotations}
Figure~\ref{fig:bar_with_ci_vs_chatbot_arena_gpt4o} shows the BT preference computed out of Arena-Lite. 
For judge, we used \texttt{gpt-4o}. As mentioned in the caption, the BT preference are bootstrapped median value from 500 trials. 95\% confidence intervals also plotted as an error bar, which look negligible in scale compared to observed values. Matches are performed over Arena-Hard-Auto benchmark dataset (500 prompts).

\subsection{Binary search vs. Win rate over baseline}
\label{appendix:Binary search vs. Win rate over baseline}
\begin{algorithm}
\caption{Binary Search for Enlisting new LLM to a leaderboard}
\begin{algorithmic}[1]
\Require Leaderboard $L$, new model $m_{\text{new}}$, test prompts $X$, outputs $O_{ij}$, assumes $|X| > |L| > n_{\text{comparisons}}$
\Ensure Updated leaderboard $L'$ with $m_{\text{new}}$ placed
\State $n_{\text{comparisons}} \gets \lfloor \log_2(|L|) \rfloor$
\State $n_{\text{matches}} \gets \lfloor |X| / n_{\text{comparisons}} \rfloor$ 

\Function{BinarySearchPlacement}{$L, m_{\text{new}}$}
  \State X $\gets$ Shuffle(X)
  \State X $\gets$ concat(X;X)
  \State low $\gets 0$
  \State high $\gets |L| - 1$
  \While{low $\leq$ high}
    \State mid $\gets \lfloor (\text{low} + \text{high}) / 2 \rfloor$
    \State wins $\gets 0$
    \For{$i \gets 1$ to $n_{\text{matches}}$}
      \State $x \gets$ $X$.pop() 
      \If{Match($m_{\text{new}}, L[\text{mid}], x) = m_{\text{new}}$}
        \State wins $\gets$ wins $+ 1$
      \EndIf
    \EndFor
    \If{wins $> n_{\text{matches}} / 2$}
      \State high $\gets$ mid $- 1$
    \ElsIf{wins $< n_{\text{matches}} / 2$}
      \State low $\gets$ mid $+ 1$
    \ElsIf{$|X|>$0} 
      \State continue \Comment{Ensure tie}       
    \Else 
      \State \Return mid, tie \Comment{Tie}
    \EndIf
  \EndWhile
  \State \Return low, non-tie \Comment{Position found}
\EndFunction

\Function{UpdateLeaderboard}{$L, m_{\text{new}}$}
  \State position, istie $\gets$ BinarySearchPlacement($L, m_{\text{new}}$)
  \State $L' \gets L$.insert(position, $m_{\text{new}}$, istie)
  \State \Return $L'$
\EndFunction
\end{algorithmic}
\label{appendix:algo:va_scenario_2}
\end{algorithm}

\subsubsection{Binary Search}
We tried binary search placement of a newly added LLM to the leaderboard without baseline output in Table~\ref{tab:binarysearch_vs_anchored-detailed}. Details of how we implemented binary search are attached in Algorithm \ref{appendix:algo:va_scenario_2}, Appendix. It turns out that binary search based on already built leaderboard ranks is not as reliable compared to utilizing the best model's outputs as a baseline. Therefore, when adding a newcomer LLM to pre-existent leaderboard, we could utilize the already submitted responses as a baseline from the 1st placed LLM.

\begin{table}[htbp]
\centering

\resizebox{\linewidth}{!}{%
\begin{tabular}{lcccc}
\hline
$\bar{|\Delta_\text{rank}|}$ ($\downarrow$)& gt=1-6& 7-13& 14-19 (20)& total avg.\\
\hline
binary search (4o)& \textbf{0.92}& 1.84& 2.13
& 1.72
\\
comp. to 1st (4o)& 1.98& \textbf{1.55}& \textbf{1.57}
& \textbf{1.39}
\\
\hline
binary search (4o-mini)& 1.27& 1.82& \textbf{1.21}
& 1.5
\\
comp. to 1st (4o-mini)& \textbf{1.00}& \textbf{1.43}& 1.43
& \textbf{1.37}
\\
\end{tabular}
}
\vspace{1em}

\caption{
Comparison of the binary search method versus using the top-performing model's response as a baseline (\textit{comp. to 1st}) for inserting a new LLM into the leaderboard. We report the mean rank deviation ($\bar{|\Delta_{\text{rank}}|}$) from the ground-truth leaderboard as an additional error metric. For further details, see Algorithm \ref{appendix:algo:va_scenario_2} in Appendix.
}
\label{tab:binarysearch_vs_anchored}
\end{table}

\subsubsection{Comparing to the most Performant Model so far: Converting Ratings Table back to Win Rates}
 Assuming we preserved a set of match results and model outputs from the last benchmarking, we could benefit from those to perform insertion. One could pick an appropriate \textit{anchor} LLM as a baseline in a leaderboard to estimate the skill of a newcomer. Using previous matches from the tournaments that built the leaderboard could be used for estimating win rates over the baseline. This is the same as converting the preference ratings table into a win rate leaderboard. Since the leaderboard is not built with full-grid matches but with tournaments, there would be some missing matches against the baseline regardless we have picked. There are two ways to estimate the win rate over the baseline model. We could just count the matches given are enough in amount, or we could also convert BT preference back to $P(i>a)$ to use it directly for scoring for the model ranks in the leaderboard. Reminding that BT preference rating is for expecting a likely outcome of the match, this should work. After this win rate of the newcomer model $P^*(n>a)=\frac{\text{count(n wins)}}{|X|}$ could be directly compared for enlisting.

\begin{table}[htbp]
\centering

\resizebox{\linewidth}{!}{%
\begin{tabular}{lcccccc|c}
\hline
$|\Delta_\text{rank}|$ ($\downarrow$) & gt=1 & 2 & 3 & 4 & 5 & 6 & avg. \\
\hline
binary search& 0.09& 1.24& \textbf{1.75}& \textbf{1.55}& 1.26& 1.10& \textbf{0.92}\\
(4o) & \tiny{(.04/-.03)}& \tiny{(.14/-.14)}& \tiny{(.09/-.09)}& \tiny{(.07/-.06)}& \tiny{(.08/-.08)}& \tiny{(.10/-.09)}&  
\\
anchored & \textbf{0.00} & \textbf{1.01} & 1.95 & 2.00 & \textbf{0.96} & \textbf{0.30} & 1.98 
\\
(4o) & \tiny{(0.00/0.00)} & \tiny{(0.01/-0.01)} & \tiny{(0.02/-0.02)} & \tiny{(0.00/0.00)} & \tiny{(0.02/-0.02)} & \tiny{(0.04/-0.04)} & 
\\
\hline
binary search& 0.52 & 0.85& 0.59& 2.03& 1.20& 2.45& 1.27 
\\
(4o-mini) & \tiny{(.09/-.07)}& \tiny{(.12/-.11)}& \tiny{(.10/-.09)}& \tiny{(.02/-.02)}& \tiny{(.05/-.05)}& \tiny{(.07/-.06)}&  
\\
anchored & 0.00 & 0.00 & 1.00 & 2.00 & 2.00 & 1.00 & \textbf{1.00}\\
(4o-mini) & \tiny{(0.00/0.00)} & \tiny{(0.00/0.00)} & \tiny{(0.00/0.00)} & \tiny{(0.00/0.00)} & \tiny{(0.00/0.00)} & \tiny{(0.00/0.00)} & \\
\hline
\end{tabular}
}
\vspace{1em}

\resizebox{\linewidth}{!}{%
\begin{tabular}{ccccccc|c}
\hline
7 & 8 & 9 & 10 & 11 & 12 & 13 & avg. \\
\hline
 1.31& 1.27& 2.22 & 1.74& 2.27& 2.23& 1.86& 1.84 
\\
 \tiny{(.10/-.10)} & \tiny{(.11/-.11)}& \tiny{(.14/-.12)}& \tiny{(.09/-.09)} & \tiny{(.12/-.11)}& \tiny{(.12/-.12)}& \tiny{(.07/-.07)} & 

\\
0.30 & 3.68 & 1.09 & 1.03 & 2.97 & 0.78 & 1.00 & \textbf{1.55}\\
\tiny{(0.04/-0.04)} & \tiny{(0.04/-0.04)} & \tiny{(0.03/-0.03)} & \tiny{(0.02/-0.01)} & \tiny{(0.02/-0.02)} & \tiny{(0.05/-0.05)} & \tiny{(0.00/0.00)} & 

\\ 
\hline
 0.69 & 0.85& 3.89& 1.95& 2.10& 2.37& 0.88& 1.82 
\\
 \tiny{(.07/-.06)}& \tiny{(.09/-.09)} & \tiny{(.12/-.11)}& \tiny{(.06/-.05)}& \tiny{(.03/-.03)}& \tiny{(.10/-.11)}& \tiny{(.12/-.11)}& 

\\ 
0.51& 0.52& 3.50& 1.00 & 1.00 & 3.00 & 0.50 & \textbf{1.43}\\
\tiny{(0.49/-0.51)}& \tiny{(0.48/-0.52)}& \tiny{(0.49/-0.51)} & \tiny{(0.00/0.00)} & \tiny{(0.00/0.00)} & \tiny{(0.00/0.00)} & \tiny{(0.50/-0.50)} & \\
\hline
\end{tabular}
}
\vspace{1em}

\resizebox{\linewidth}{!}{%
\begin{tabular}{ccccccc|c}
\hline
14 & 15 & 16 & 17 & 18 & 19 & 20 & avg. \\
\hline
 1.40& 3.07& 0.80& 1.47& 5.00& 0.96 & - & 2.13 
\\
 \tiny{(.04/-.05)} & \tiny{(.11/-.11)} & \tiny{(.08/-.09)} & \tiny{(.05/-.04)} & \tiny{(.11/-.11)} & \tiny{(.08/-.09)} & & 
\\
2.00 & 2.00 & 1.00 & 1.21 & 3.00 & 0.21 & - & \textbf{1.57}\\ 
\tiny{(0.00/0.00)} & \tiny{(0.00/0.00)} & \tiny{(0.00/0.00)} & \tiny{(0.03/-0.04)} & \tiny{(0.00/0.00)} & \tiny{(0.04/-0.03)} & & 
\\
\hline
 1.45& 4.20& 0.19& 0.08& 1.09& 1.08& 0.40& \textbf{1.21}\\ 
 \tiny{(.07/-.08)}& \tiny{(.17/-.17)}& \tiny{(.07/-.06)}& \tiny{(.03/-.02)} & \tiny{(.05/-.05)} & \tiny{(.05/-.05)}& \tiny{(.07/-.07)} & 
\\
1.00 & 2.00 & 2.00 & 1.00 & 1.00 & 3.00 & 0.00 & 1.43 
\\
\tiny{(0.00/0.00)} & \tiny{(0.00/0.00)} & \tiny{(0.00/0.00)} & \tiny{(0.00/0.00)} & \tiny{(0.00/0.00)} & \tiny{(0.00/0.00)} & \tiny{(0.00/0.00)} & \\
\hline
\end{tabular}
}

\caption{Binary search vs. \textit{Anchored comparison}: Mean rank deviation ($|\Delta_\text{rank}|$) from ground-truth leaderboard. Result of binary search placement and anchored comparison insert by \texttt{gpt-4o[-mini]} judge are provided with bootstrapped 95\% confidence interval (500 trials, 1000 samples, $|X|$=500, Arena-Hard-Auto~\cite{arena-hard}).}
\label{tab:binarysearch_vs_anchored-detailed}
\end{table}

\subsection{Separability In terms of Confidence Interval}
\begin{figure}
   \centering
   \includegraphics[width=1\linewidth]{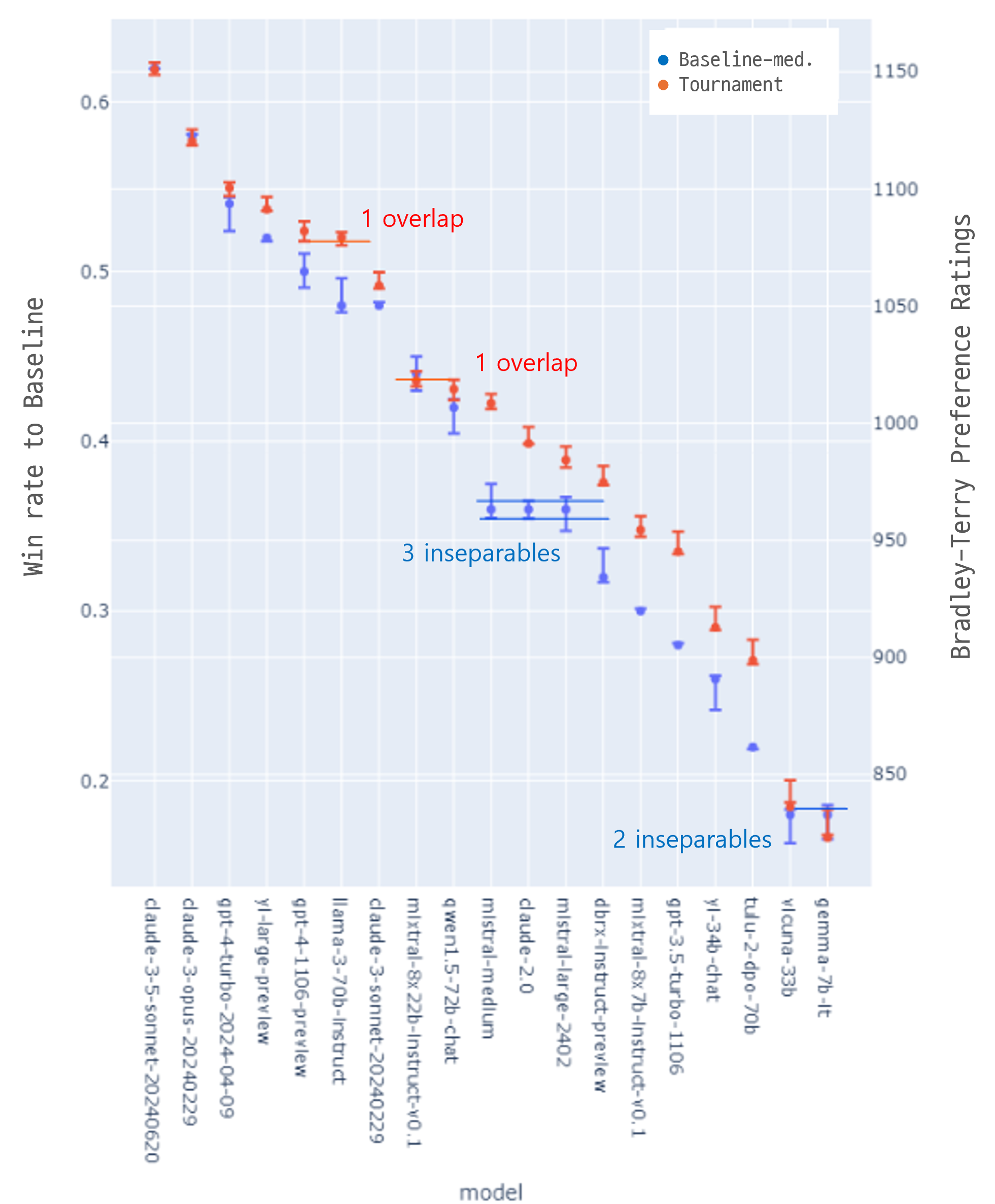}
   \caption{\texttt{gpt-4o} result of \textit{anchored comparison} and tournament approach. 1000 bootstrapped median from 500 observations used for confidence interval estimation. }
   \label{fig:gpt-4o-scatter}
 \end{figure}

To see how well the two benchmarking approach (\textit{anchored comparison} and tournament approach) separates LLMs in adjacent ranks, we provide scatter plot of Elo rating and win rate paired with error bar (95\% confidence interval). We present the both results of using \texttt{gpt-4o} (Figure~\ref{fig:gpt-4o-scatter}) and \texttt{gpt-4o-mini} (Figure~\ref{fig:gpt-4o-scatter}) as a judge. Inside the each plot, inseparables indicates the cases where any pair of datapoint co-includes each other within their range of error bars, and overlap means a certain datapoint is within some other's range of error, when it is one-sided.

 \begin{figure}
   \centering
   \includegraphics[width=1\linewidth]{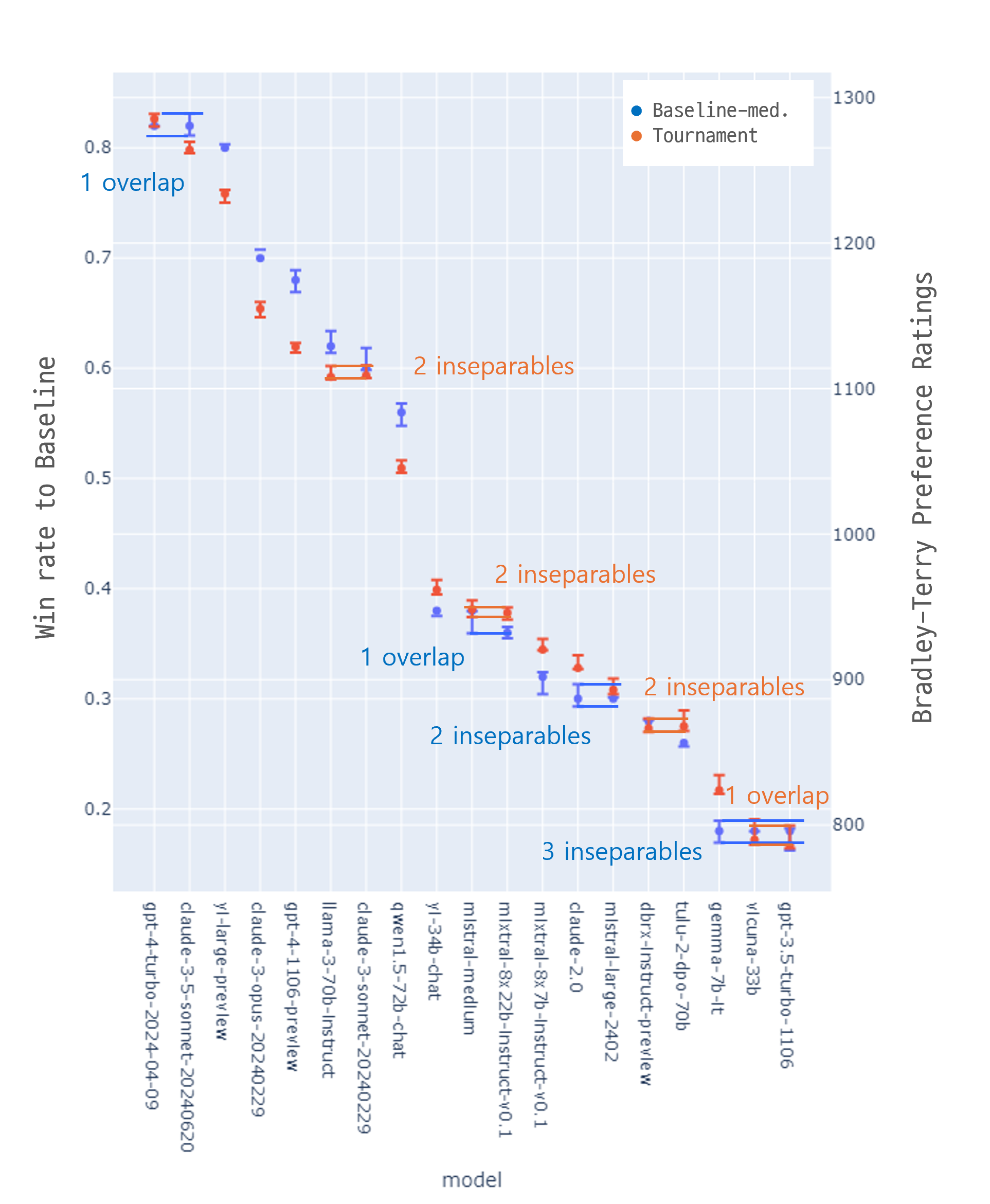}
   \caption{\texttt{gpt-4o} result of \textit{anchored comparison} and tournament approach. 1000 bootstrapped median from 500 observations used for confidence interval estimation. }
   \label{fig:gpt-4o-mini-scatter}
 \end{figure}

\subsection{Judge configuration}
\label{appendix:Prompts and Judge configuration}
\subsubsection{Evaluation Prompt}
We use the prompt from LLMBar. The prompt depicted in Figure~\ref{fig:prompt}. We added 4 questions for criteria of our own to \texttt{Metrics.txt} prompt of \cite{LLMBar}. You can refer to the original prompt in LLMBar github. 
\subsubsection{Decoding Parameters}
We did not configure decoding parameters of judge LLMs (\texttt{gpt-4o[-mini]}), which its temperature defaults to 1. The only parameter we have adjusted is maximum number of tokens to be generated, which for our prompt is less than 6 (i.e. The output of our prompt is \texttt{(a)} or \texttt{(b)}). To avoid position bias, we alternated the position of the responses from a certain model across the benchmark prompt.

\onecolumn
PROMPTS = [ \textit{\# metrics.txt from LLMBar}\\
  \{ \\
    "role": "system",
    "content": "You are a helpful assistant in evaluating the quality of the outputs for a given instruction. Your goal is to select the best output for the given instruction.", \\
  \}, \\
  \{ \\
    "role": "user",
    "content": """Select the Output (a) or Output (b) that is better for the given instruction. The two outputs are generated by two different AI chatbots respectively. \\
\\
Here are some rules of the evaluation: \\
(1) You should prioritize evaluating whether the output honestly/precisely/closely executes the instruction, then consider its helpfulness, accuracy, level of detail, harmlessness, etc. \\
(2) Outputs should NOT contain more/less than what the instruction asks for, as such outputs do NOT precisely execute the instruction. \\
(3) You should avoid any potential bias and your judgment should be as objective as possible. For example, the order in which the outputs were presented should NOT affect your judgment, as Output (a) and Output (b) are **equally likely** to be the better. \\
\\
Do NOT provide any explanation for your choice. \\
Do NOT say both / neither are good. \\
You should answer using ONLY "Output (a)" or "Output (b)". Do NOT output any other words. \\
\\
\# Instruction: \\
{instruction} \\
\\
\# Output (a): \\
{response\_a} \\
\\
\# Output (b): \\
{response\_b} \\
\\
\# Questions about Outputs:\\
Here are at most three questions about the outputs, which are presented from most important to least important. You can do the evaluation based on thinking about all the questions.\\
- Does the output well satisfy the intent of the user request?\\
- If applicable, is the output well-grounded in the given context information?\\
- Does the output itself satisfy the requirements of good writing in terms of:\\
  1) Coherence\\
  2) Logicality \\
  3) Plausibility \\
  4) Interestingness \\
\\
\\
\# Which is better, Output (a) or Output (b)? Your response should be either "Output (a)" or "Output (b)":""",\\
  \}, \\
] \textit{\# prompt ends here}\\
\\

LLMBar prompt of our use. We used \texttt{metric} variant suggested in original LLMBar paper. More preset prompts are in our Arena-Lite Demo and source (\href{https://huggingface.co/spaces/NCSOFT/ArenaLite}{\url{https://huggingface.co/spaces/NCSOFT/ArenaLite}})
\label{fig:prompt}

\end{document}